# Vector-space Analysis of Belief-state Approximation for POMDPs

**Pascal Poupart**
Department of Computer Science
University of Toronto
Toronto, ON M5S 3H5
*ppoupart@cs.toronto.edu*

**Craig Boutilier**
Department of Computer Science
University of Toronto
Toronto, ON M5S 3H5
*cebly@cs.toronto.edu*

## Abstract

We propose a new approach to value-directed belief state approximation for POMDPs. The value-directed model allows one to choose approximation methods for belief state monitoring that have a small impact on decision quality. Using a vector space analysis of the problem, we devise two new search procedures for selecting an approximation scheme that have much better computational properties than existing methods. Though these provide looser error bounds, we show empirically that they have a similar impact on decision quality in practice, and run up to two orders of magnitude more quickly.

## 1 Introduction

Partially observable Markov decision processes (POMDPs) have attracted considerable attention as a model for decision-theoretic planning. Their generality allows one to seamlessly model sensor and action uncertainty, uncertainty in the state of knowledge, and multiple objectives [1, 5]. Their computational intractability has, however, limited their practical applicability [11, 13].

An important approach to POMDPs involves constructing a value function for a *belief state MDP* offline, and maintaining a *belief state* (or distribution over system states) online, which is used to implement an optimal policy [18]. A number of approaches attacking the offline computational problems have been studied, including improved algorithms [6], the use of factored representations [2, 8], as well as numerous approximation schemes [9]. Little work has focused on the online belief state monitoring problem. Because planning state spaces grow exponentially with the number of variables, maintaining an explicit distribution over states is generally impractical. Even when concise representations such as dynamic Bayes nets (DBNs) are used, monitoring is generally intractable, since the independencies exploited by DBNs vanish over time. Boyen and Koller [3] proposed *projection schemes* for approximate monitoring, essentially breaking weaker correlations among variables to ensure tractability. Poupart and Boutilier [15] proposed *value-directed* methods for approximation, allowing the anticipated loss in expected utility guide the choice of approximation scheme.

In this paper we pursue the value-directed approach since its emphasis on minimizing impact on decision quality is a critical factor in devising useful approximations. We use the value function itself to determine which correlations can be "safely" ignored when monitoring one's belief state. We propose an alternative approach to choosing approximation schemes for monitoring in POMDPs that overcomes many of the computational bottlenecks of [15]. We introduce a *vector space formulation* of the approximation problem that allows one to construct approximation schemes with looser error bounds, but much more quickly. Despite the looser bounds, we show empirically that decision quality is rarely worse than that obtained using the more intensive approaches. Our methods work in time roughly on order of the time taken to solve a POMDP, and since they run offline, they can be used with any POMDP technique that can currently be applied. Furthermore, these methods take advantage of the factored (DBN) representations to avoid state enumeration. The *offline* cost allows much faster (approximate) *online* policy implementation. Even in cases where a POMDP must be solved in a traditional "flat" fashion, we typically have the luxury of compiling a value function offline. Thus, even for large POMDPs, we might reasonably expect to have value function information (either exact or approximate) available to direct the monitoring process. The fact that one is able to produce a value function offline does not imply the ability to monitor the process exactly in a timely online fashion.[1] Finally, our model offers a novel view of the approximation problem for belief state monitoring for POMDPs.

We briefly overview POMDPs and value-directed approximation in Section 2. We present our vector space formulation in Section 3 and provide some suggestive empirical

---
[1]While techniques exist for generating finite-state controllers for POMDPs, there are still reasons for wanting to use value-function-based approaches [14].

results in Section 4.

## 2 POMDPs and Belief State Monitoring

The key components of a POMDP are: a finite state space $\mathcal{S}$; a finite action space $\mathcal{A}$; a finite observation space $\mathcal{Z}$; and a reward function $R : \mathcal{S} \rightarrow \mathbf{R}$. Actions induce stochastic state transitions with specified probabilities, and an agent is provided with noisy observations of the system state (with specified probabilities). A reward is received at each state and an agent's objective is to control the system through judicious choice of action to maximize the expected reward obtained over some horizon of interest.

The rewards obtained over time by an agent adopting a specific course of action can be viewed as random variables $R^{(t)}$. Our aim is to construct a *policy* that maximizes the expected sum of discounted rewards $E(\sum_{t=0}^{\infty} \gamma^t R^{(t)})$ (where $\gamma$ is a discount factor less than one). An optimal course of action can be determined by considering the fully observable *belief state MDP*, where *belief states* (distributions over $\mathcal{S}$) form states, and a policy $\pi : \mathcal{B} \rightarrow \mathcal{A}$ maps belief states into action choices. A key result of Sondik [18] showed that the value function $V$ for a finite-horizon problem is piecewise-linear and convex and can be represented as a finite collection of $\alpha$-vectors; for infinite-horizon problems, a finite collection generally offers a good approximation. Specifically, one can generate a collection $\aleph$ of $\alpha$-vectors, each of dimension $|\mathcal{S}|$, such that $V(b) = \max_{\alpha \in \aleph} b \cdot \alpha$. In Figure 1 the value function is given by the upper surface of the five vectors shown. Each vector is associated with a specific (course of) action. For finite horizon POMDPs, a set $\aleph^k$ is generated for each stage $k$ of the process. Algorithms exist that construct efficient representations of $\alpha$-vectors, such as decision trees or *algebraic decision diagrams (ADDs)*, when the POMDP is specified concisely using DBNs [2, 8].

Insight into the nature of POMDP value functions can be gained by examining Monahan's [12] method for solving POMDPs. Monahan's algorithm proceeds by producing a sequence of $k$-stage-to-go value functions $V^k$, each represented by a set of $\alpha$-vectors $\aleph^k$. Each $\alpha \in \aleph^k$ denotes the value (as a function of the belief state) of executing a $k$-step *conditional plan*. More precisely, let the $k$-step *observation strategies* be the set $OS^k$ of mappings $\sigma : \mathcal{Z} \rightarrow \aleph^{k-1}$. Then each $\alpha$-vector in $\aleph^k$ corresponds to the value of executing some action $a$ followed by implementing some $\sigma \in OS^k$; that is, it is the value of doing $a$, and executing the $k - 1$-step plan associated with the $\alpha$-vector $\sigma(z)$ if $z$ is observed. Using $CP(\alpha)$ to denote this plan, we have that $CP(\alpha) = \langle a; \text{if } z_i, CP(\sigma(z_i)) \forall z_i \rangle$. We informally write this as $\langle a; \sigma \rangle$. We write $\alpha(\langle a; \sigma \rangle)$ to denote the $\alpha$-vector reflecting the value of this plan.

The implementation of a policy requires that one monitor belief state $b$ over time so that it may be "plugged" into the value function (or $\aleph$) to make a suitable action choice. Be-lief states can be maintained by standard Bayesian methods; but when $|\mathcal{S}|$ is large, the cost is prohibitive. This is especially true when $\mathcal{S}$ is determined by a set of variables $\mathbf{X}$ (and $|\mathcal{S}| = O(2^{|\mathbf{X}|})$). In such cases, DBNs can be used to represent the dynamics of POMDPs and DBN inference techniques that exploit conditional independence among variables can be applied to make monitoring more efficient. Unfortunately, as shown by Boyen and Koller [3], in many problems most if not all variables of DBNs tend to become correlated over time so DBNs offer no significant savings.

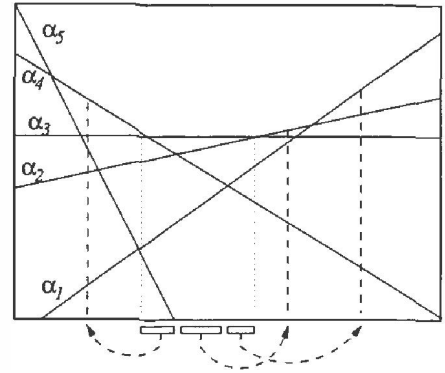

Figure 1: The Switch Set $Sw(\alpha_3)$ of $\alpha_3$

Boyen and Koller introduced *projection schemes* as a method to approximate belief states. Given variables $\mathbf{X}$ defining $\mathcal{S}$, a projection is a set $S$ of subsets of $\mathbf{X}$ with each variable in at least one subset. Correlations among variables within a subset are preserved while the subsets are assumed to be independent. For instance, if $\mathbf{X} = \{A, B, C\}$, then projection $S = \{AB, C\}$ approximates the exact belief state $b = \Pr(A, B, C)$ with $b' = \Pr(AB) \Pr(C)$. The assumed independence allows more efficient monitoring using DBNs: at most, one maintains marginals over each subset in $S$.

The choice of projection scheme (or any other approximation) can have a dramatic impact on decision quality in a POMDP, since the approximate belief $b'$ can lead to the choice of a suboptimal course of action. Poupart and Boutilier [15] propose a *value-directed approximation* framework allowing computation of bounds on the loss in expected utility for projection schemes, and search methods for choosing projections that tradeoff decision quality with monitoring efficiency. The techniques are computationally intensive (potentially requiring time quadratic in the solution time of the POMDP); but this *offline* computation produces a projection scheme that improves *online* monitoring efficiency with minimal sacrifice in decision quality. We briefly outline this model.

Assume a POMDP has been solved giving the set $\aleph$ of $\alpha$-vectors with $\alpha \in \aleph$. Let $R(\alpha)$ be the *optimal region* for $\alpha$ (i.e., the set of belief states $b$ such that $\alpha$ is maximal for $b$). Given a projection scheme $S$, the *switch set* $Sw(\alpha)$ is the set of $\alpha'$ such that $S(b) \in R(\alpha')$ for some $b \in R(\alpha)$. Thus, $S$ could induce one to believe $\alpha'$ has maximum value at the current belief state instead of $\alpha$, thereby erroneously "switching to" the plan corresponding to $\alpha'$ from $\alpha$ by using $S$. Figure 1 illustrates a switch set $Sw(\alpha_3) = \{\alpha_1, \alpha_2, \alpha_4\}$. Switch sets can be computed by solving a nonlinear program for each $\alpha \in \aleph$. Linear programs (LPs) can be used to more effectively produce a superset of the switch set [15].

Given the switch sets (or supersets thereof), one can compute an upper bound $B_S^k$ on the loss in expected value for a single approximation using $S$ at $k$ stages to go:

$$B_S^k = \max_{\alpha \in \aleph^k} \max_b \max_{\alpha' \in Sw_S^k(\alpha)} b \cdot (\alpha - \alpha')$$

When multistage approximations are applied, one can devise an *alternative set* which is similar in spirit to the switch set. The alternative set $Alt(\alpha)$ is the set of all $\alpha$-vectors corresponding to alternative plans that may be executed as a result of repeatedly approximating the belief state at all future time steps (see [15] for a precise definition). $Alt(\alpha)$ is constructed with a dynamic programming procedure similar to incremental pruning [6]. One can define an upper bound $E_S^k$ on the loss in expected value due to *successive* belief state approximations using $S$ for $k$ stages to go:

$$E_S^k = \max_{\alpha \in \aleph^k} \max_b \max_{\alpha' \in Alt_S^k(\alpha)} b \cdot (\alpha - \alpha')$$

These bounds can be extended to infinite-horizon problems. Given the bounds $B$ and $E$, one can search for an "optimal" projection scheme by looking for the projection that minimizes one of those bounds. The space of projection schemes is very large (factorial in the number of variables), but exhibits a nice lattice structure. Figure 2 illustrates the lattice of projection schemes when the state space is defined by the joint instantiation of variables $A$, $B$ and $C$. Each point denotes a projection scheme, with "descendents" of any projection corresponding to more coarse-grained projections. As we move down the lattice, accuracy increases since the number of correlations among the variables preserved in our belief state is increased (hence, error bounds $B$ and $E$ monotonically decrease); but monitoring efficiency decreases as we move downward for the same reason. A number of search procedures can be used to traverse the lattice, using the error bounds to guide the search. For example, a simple (and incremental) greedy scheme is proposed in [15]. The search is stopped when a suitable accuracy/efficiency tradeoff has been reached.

## 3 Vector Space Analysis

We now provide a vector space analysis of belief state approximation by projection, showing in Section 3.1 that projections allow movement of belief state only in certain directions (defining a subspace). This allows us to view $\alpha$-vectors as determining gradients of value in different directions: approximations whose directions give *similar* value gradients are less likely to cause switching (hence minimizing error). In Section 3.2 we use this to design faster switch test algorithms than those described above, though yielding looser bounds. In Section 3.3 we devise a new vector-space search algorithm to find projections without directly trying to minimize these error bounds, instead relying on value gradient similarity.

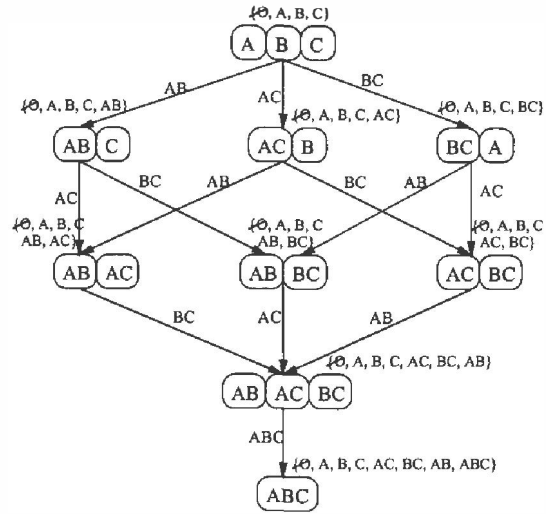

Figure 2: Lattice of Projection Schemes

### 3.1 Vector space formulation

Given a projection $S$ over $\mathbf{X}$, let $b$ and $b' = S(b)$ be points in belief space. Define $d = b' - b$ to be the *displacement vector* from $b$ to $b'$. Projection $S$ determines a set of linear equations constraining $b$ in terms of $b'$. For example, if $\mathbf{X} = \{X, Y\}$ and $S = \{X, Y\}$ (i.e., $S$ treats $X, Y$ as independent), we have:

$$d(xy) + d(x\bar{y}) + d(\bar{x}y) + d(\bar{x}\bar{y}) = 0$$
$$d(xy) + d(x\bar{y}) = 0$$
$$d(xy) + d(\bar{x}y) = 0$$

Geometrically, we interpret each equation as a hyperplane; and their intersection (or solution space) is a line through the origin representing a one-dimensional (in this example) *subspace*. This subspace captures the set of all displacement vectors resulting from the application of $S$ (w.r.t. $b'$). Since all possible displacement vectors lie on the same line, they must all have the same direction (vectors with opposite orientation are assumed to have the same direction).

To illustrate, let $b(x) = 0.3$ and $b(y) = 0.4$. The approximate belief state using $S$ above gives:

$$\begin{aligned} b'(xy) &= b(x)b(y) &= 0.12 \\ b'(x\bar{y}) &= b(x)b(\bar{y}) &= 0.18 \\ b'(\bar{x}y) &= b(\bar{x})b(y) &= 0.28 \\ b'(\bar{x}\bar{y}) &= b(\bar{x})b(\bar{y}) &= 0.42 \end{aligned}$$

Figure 3 shows a three-dimensional belief space for belief states $xy$, $x\bar{y}$, $\bar{x}y$ and $\bar{x}\bar{y}$.[2] All belief states $b$ with $b(x) =$

---
[2] We omit dimension $b(\bar{x}\bar{y})$ as probabilities sum to 1.

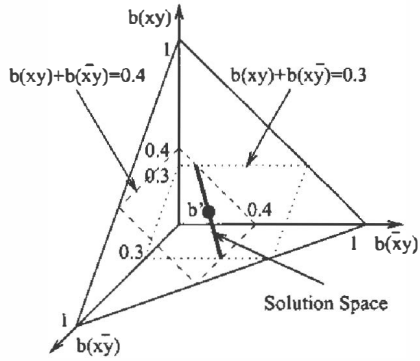

Figure 3: Solution space of possible exact belief states $b$

$$\begin{aligned}
\max \quad & x \\
\text{s.t.} \quad & b \cdot (\alpha_i - \alpha_j) \geq x \\
& b' \cdot (\alpha_j - \alpha_i) \geq x \\
& b'(m') = b(m') \quad \forall m' \subseteq m, \forall m \in S \\
& \sum_s b(s) = 1 \\
& b(s) \geq 0 \quad \forall s \\
& b'(s) \geq 0 \quad \forall s
\end{aligned}$$

Table 1: Linear VS-switch test for projection schemes. This LP has a strictly positive objective value iff there is some $b \in R(\alpha_i)$ and $b' \in R(\alpha_j)$ such that $b(m') = b'(m')$ for any subset $m'$ of variables contained in some marginal $m \in S$.

$0.3$ lie in a hyperplane, and similarly for $b(y) = 0.4$. Their intersection is the set $\{b : b' = S(b)\}$, and all displacement vectors for $b'$ have the same direction. (For marginals other than $0.3$ and $0.4$, the hyperplanes and their intersection shift, but remain parallel).

Let $D_S$ be the *displacement subspace* spanned by the set of all displacement vectors induced by $S$: it is completely characterized by its marginals (elements) and it describes the directions of all displacements. In general, $D_S$ is a $(2^{|X|} - c)$-dimensional subspace, where $c$ is the number of constraints, since it is the solution space of $c$ linearly independent equations, each corresponding to a constraint $d(m) = 0$. ($c$ is the number of subsets of variables contained in some subset $m \in S$, as above.) This is obvious when we rewrite the constraints as $v_m \cdot d = 0$, where $v_m$ is a boolean $|S|$-vector with 1 at states with all $X \in m$ true and 0 at states with some $X \in m$ false.[3] In our example, we have:

$$\begin{array}{cccccc}
& & xy & x\bar{y} & \bar{x}y & \bar{x}\bar{y} \\
v_\emptyset & = & ( \ 1 & 1 & 1 & 1 \ ) \\
v_X & = & ( \ 1 & 1 & 0 & 0 \ ) \\
v_Y & = & ( \ 1 & 0 & 1 & 0 \ )
\end{array}$$

Let $D_S^\perp$ be the subspace spanned by the vectors $v_m, m \in S$; the space $D_S^\perp$ is the *null space* of $D_S$ (i.e., the set of vectors perpendicular to each vector in $D_S$).

### 3.2 Vector space switch test

We will see below that the subspaces $D_S$ and $D_S^\perp$ allow a nice characterization of a new switch test. We first consider a simple relaxation of the switch test of [15]. Recall from Section 2 that approximation $S$ could induce an agent to switch from optimal vector $\alpha_i$ to suboptimal vector $\alpha_j$ if $S(b) \in R(\alpha_j)$ for some $b \in R(\alpha_i)$. The idea behind the relaxed *vector space (VS) switch test* is to simply apply the same technique *ignoring the presence of other $\alpha$-vectors*. The VS switch test asks whether there is some belief state $b$ for which $b \cdot \alpha_i > b \cdot \alpha_j$ yet $S(b) \cdot \alpha_i < S(b) \cdot \alpha_j$. If so, we say $\alpha_j$ is in the *VS-switch set* of $\alpha_i$. This is equivalent to asking if $\alpha_j \in Sw(\alpha_i)$ when all vectors except these two are removed from $\aleph$. Note that the VS-switch set is a superset of the true switch set.

---
[3]The generalization to nonboolean variables is straightforward.

Since the constraints relating $b$ and $S(b)$ are nonlinear, VS-switch sets can be computed using nonlinear programs. We can define a simpler *linear* VS-switch test as in Table 1 which produces a superset of the VS-switch set. This LP is a relaxation of the LP switch test [15].

Now define $\alpha_{ij} = \alpha_i - \alpha_j$ to be a vector representing the *difference* in expected value for executing $\alpha_j$ instead of $\alpha_i$. We can show that the VS-switch test for $\alpha_i$ and $\alpha_j$ is positive iff $\alpha_{ij} \notin D_S^\perp$. Consider $\alpha_{ij}$ as a *gradient* that measures the error induced by an approximation when it causes a switch from $\alpha_i$ to $\alpha_j$. After an approximation, if this difference changes considerably, the agent is likely to choose the wrong maximizing $\alpha$-vector. Define the *relative error*, $\delta_{ij}$, of this change in the relative assessment of $\alpha_i$ with respect to $\alpha_j$ as:

$$\begin{aligned}
\delta_{ij} & = b(\alpha_i - \alpha_j) - S(b)(\alpha_i - \alpha_j) \\
& = d \cdot \alpha_{ij}
\end{aligned}$$

Here $\alpha_{ij}$ can be viewed as a gradient since approximations corresponding to displacement vectors $d$ parallel to $\alpha_{ij}$ maximize the magnitude of $d \cdot \alpha_{ij}$. In general, the angle between $d$ and $\alpha_{ij}$ is a good indicator of approximation error. In particular, if they are perpendicular, their dot product is zero and the relative assessment of $\alpha_i$ and $\alpha_j$ remains unchanged, preventing any switch. By definition, the subspace $D_S^\perp$ is the set of vectors perpendicular to all displacement vectors possibly induced by $S$, so when $\alpha_{ij}$ is a member of $D_S^\perp$, all possible displacement vectors are perpendicular to $\alpha_{ij}$ and consequently there cannot be a switch from $\alpha_i$ to $\alpha_j$. Thus $\alpha_{ij} \notin D_S^\perp$ iff the VS-switch test is positive.

This fact provides for a much more efficient method to compute switch sets than the LP of Table 1. We decompose $\alpha_{ij}$ in two orthogonal vectors corresponding to the projections of $\alpha_{ij}$ onto $D_S^\perp$ and $D_S$:

$$\alpha_{ij} = proj(\alpha_{ij}, D_S^\perp) + proj(\alpha_{ij}, D_S)$$

(where $proj(\alpha, D)$ stands for the projection of $\alpha$ onto $D$). If $\alpha_{ij} \in D_S^\perp$, then $proj(\alpha_{ij}, D_S^\perp) = \alpha_{ij}$ and, consequently, $proj(\alpha_{ij}, D_S)$ is the zero-vector; otherwise, $proj(\alpha_{ij}, D_S)$ is nonzero. We can thus determine if $\alpha_{ij} \in D_S^\perp$ by measuring the length of $proj(\alpha_{ij}, D_S)$. We have that $\|proj(\alpha_{ij}, D_S)\|_2 = 0$ when $\alpha_{ij} \in D_S^\perp$, and $\|proj(\alpha_{ij}, D_S)\|_2 > 0$ when $\alpha_{ij} \notin D_S^\perp$. In particular, the squared length of $proj(\alpha_{ij}, D_S)$ can be computed by the following equation:

$$\|proj(\alpha_{ij}, D_S)\|_2^2 = \alpha_{ij} \cdot \alpha_{ij} - \sum_{v \in \mathcal{D}_S^\perp} (\alpha_{ij} \cdot v)^2 \quad (1)$$

Here $\mathcal{D}_S^\perp$ is some orthonormal basis spanning $D_S^\perp$. The spanning set of vectors $v_m$ above can be used to generate several orthonormal bases using the Gram-Schmidt orthogonalization process and normalizing. We consider a specific orthonormal basis in particular—which we refer to as $\mathcal{D}_S^\perp$—because of its factored representation. For problems involving binary variables, every vector in $\mathcal{D}_S^\perp$ consists of a sequence of 1's and $-1$'s (before normalization). The unnormalized basis vector $\bar{v}_m$ associated with subset $m$ has a 1 in every component corresponding to a state with an even number of true variables in $m$ and $-1$ in every component corresponding to a state with an odd number of true variables in $m$. For instance, projection $S = \{XY, YZ\}$ has six marginals ($\emptyset, X, Y, Z, XY$ and $YZ$), yielding the following basis vectors:[4]

|  | $xyz$ | $xy\bar{z}$ | $x\bar{y}z$ | $x\bar{y}\bar{z}$ | $\bar{x}yz$ | $\bar{x}y\bar{z}$ | $\bar{x}\bar{y}z$ | $\bar{x}\bar{y}\bar{z}$ |  |
|---|---|---|---|---|---|---|---|---|---|
| $\bar{v}_\emptyset = ($ | 1 | 1 | 1 | 1 | 1 | 1 | 1 | 1 | $) / \sqrt{|S|}$ |
| $\bar{v}_X = ($ | $-1$ | $-1$ | $-1$ | $-1$ | 1 | 1 | 1 | 1 | $) / \sqrt{|S|}$ |
| $\bar{v}_Y = ($ | $-1$ | $-1$ | 1 | 1 | $-1$ | $-1$ | 1 | 1 | $) / \sqrt{|S|}$ |
| $\bar{v}_Z = ($ | $-1$ | 1 | $-1$ | 1 | $-1$ | 1 | $-1$ | 1 | $) / \sqrt{|S|}$ |
| $\bar{v}_{XY} = ($ | 1 | 1 | $-1$ | $-1$ | $-1$ | $-1$ | 1 | 1 | $) / \sqrt{|S|}$ |
| $\bar{v}_{YZ} = ($ | 1 | $-1$ | $-1$ | 1 | 1 | $-1$ | $-1$ | 1 | $) / \sqrt{|S|}$ |

With this orthonormal basis, we can implement VS-switch tests very effectively, without recourse to the LP in Table 1. We must simply compute Eq. 1 which requires $O(c)$ dot products. If unstructured, each dot product requires $O(|S|)$ elementary operations, for a total time of $O(c|S|)$. The use of factored representations such as ADDs considerably improves this running time. Each basis vector has only two distinct values, and yields a very compact ADD representation. Assuming that the POMDP has been solved to produce ADD representations of the $\alpha$-vectors, then the $\alpha_{ij}$ will have compact representations, and the dot products will be computed very efficiently: often a small constant independent of the size of the state space. Hence, for sufficiently structured POMDPs, the effective running time of a VS-switch test is $O(c)$.

By comparison, solving the linear program of an LP-switch test [15] is polynomial in the number of constraints $c$ and the size of the state space. Furthermore, ADDs do not provide as useful a speed up for LPs since the effective state space is the intersection of the abstract state space of all the constraints. The price paid is that the $B$ and $E$ bounds computed using the VS-switch test will generally be looser than that using the original LP test. As in Section 2, these bounds can be used to search the lattice of projection schemes for making appropriate time-decision quality tradeoffs.

### 3.3 Vector space search

In this section we describe an alternative search method based on the relative error expression $\delta_{ij}$. We do not compute switch sets at all, nor attempt to minimize worst-case error bounds as above. This new *vector-space (VS) search* process instead seeks a projection $S$ which defines a displacement subspace $D_S$ that is as perpendicular as possible to all gradients $\alpha_{ij}$. This is motivated by the observation that the more perpendicular the direction of an approximation with respect to $\alpha_{ij}$, the smaller the magnitude of $\delta_{ij}$ and, consequently, the less likely a switch will occur. Technically, this is done by minimizing the squared length of the projection of each gradient $\alpha_{ij}$ on $D_S$ (as in Eq. 1).

The length of $proj(\alpha_{ij}, D_S)$ has a special interpretation: it corresponds to the greatest (absolute) *relative error rate* for an approximation in some direction $d \in D_S$. The relative error rate corresponding to displacement vector $d$ is the relative error induced by a *unit* displacement in the direction of $d$:

$$\frac{d}{\|d\|_2} \cdot \alpha_{ij}$$

Hence, by choosing a projection $S$ that minimizes Eq. 1, we are minimizing the (squared) worst relative error rate that may result from projection $S$. When ignoring the distance between the exact and approximate belief states, the relative error rate permits us to quantify how bad an approximation in some direction is likely to be. Each projection $S$ constrains approximations to directions within the subspace $D_S$. The direction $d \in D_S$ with the highest (absolute) relative error rate has this worst relative error rate, which also happens to be $\|proj(\alpha_{ij}, D_S)\|_2$. Thus, it is desirable to try to minimize Expression 1.

Ideally we should choose an $S$ that simultaneously minimizes Eq. 1 for every gradient $\alpha_{ij}$ ($j \neq i$). In the absence of any prior information about the relative importance of each gradient, we suggest two simple schemes: (a) minimize the sum of squared lengths of each projection; or (b) minimize the squared length of the greatest projection:

$$\sum_{j \neq i} \|proj(\alpha_{ij}, D_S)\|_2^2$$
$$= \sum_{j \neq i} (\alpha_{ij} \cdot \alpha_{ij} - \sum_{v \in D_S^\perp} v \cdot \alpha_{ij}) \quad (2)$$
$$\max_{j \neq i} \|proj(\alpha_{ij}, D_S)\|_2^2$$
$$= \max_{j \neq i} (\alpha_{ij} \cdot \alpha_{ij} - \sum_{v \in D_S^\perp} v \cdot \alpha_{ij}) \quad (3)$$

---
[4]This definition can be generalized to non-binary variables.

We refer to these schemes as the *sum* and the *max* error estimators, respectively, for projection schemes. Of course, many other schemes could be proposed.

Given a vector $\alpha_i \in \aleph$, VS search uses either Eq. 2 or Eq. 3 above to find a good projection $S$ as follows. Starting at the root, we traverse the lattice of projection schemes (Figure 2) downward in a greedy manner. At each node, we pick the most promising child by minimizing Eq. 2 or Eq. 3 The computational complexity of a VS search is fairly low as it avoids LPs. Its running time is $O(nc^3|\aleph|^2|\mathcal{S}|)$, since one good projection must be found for each of the $|\aleph|$ regions $R(\alpha)$. For each region, $O(nc^2)$ nodes in the lattice are traversed, each requiring the evaluation of Eq. 2 or Eq. 3 which both take $O(c|\aleph||\mathcal{S}|)$ elementary operations.

The VS search can also be streamlined. The constraints of a node $S$ are essentially the same as the constraints of its parent node $S'$ with one extra constraint corresponding to the marginal $m$ that labels the edge connecting the two nodes. Since there is one basis vector per constraint, the following equation holds:

$$\mathcal{D}_S^\perp = \mathcal{D}_{S'}^\perp \cup \{\bar{v}_m\}$$

This means that both Eq. 2 and Eq. 3 can be computed incrementally as the lattice is traversed downward:

$$\sum_{j \neq i} \|proj(\alpha_{ij}, D_S)\|_2^2$$
$$= \sum_{j \neq i} \|proj(\alpha_{ij}, D_{S'})\|_2^2 - \bar{v}_m \cdot \alpha_{ij}$$
$$\max_{j \neq i} \|proj(\alpha_{ij}, D_S)\|_2^2$$
$$= \max_{j \neq i} \|proj(\alpha_{ij}, D_{S'})\|_2^2 - \bar{v}_m \cdot \alpha_{ij}$$

This incremental computation scheme for traversing the lattice reduces the running time to $O(nc^2|\aleph|^2|\mathcal{S}|)$ since only one dot product needs to be computed instead of one for each of the $c$ constraints. This running time is significantly smaller than $O(nc^{2+k}|\aleph||\mathcal{S}|^k)$ for the $B$-bound or $E$-bound greedy search with LP-switch tests used in [15]. As for the $B$-bound or $E$-bound greedy search with VS-switch tests, the running time $O(nc^3|\aleph||\mathcal{S}|)$ is comparable. The VS search has an extra $|\aleph|$ factor, but one less $c$ factor. In practice, $|\aleph|$ is usually larger than $c$, so the VS search is actually slower. Again, the upper bounds on running times are given in terms of $|\mathcal{S}|$, but in practice, factored representations can drastically reduce the size of the effective state space for structured POMDPs.

## 4 Empirical Evaluation

Three test problems were used to carry out the experiments. The first POMDP is essentially the coffee problem introduced by Boutilier and Poole [2]. The second POMDP is a variation of the widget problem described by Draper, Hanks and Weld [7]. The third POMDP is inspired from the pavement maintenance problem described by Puterman [17]. Since the analysis of the experiments doesn't require any specific domain knowledge, the reader is referred to [14] in which the full specification of those problems is given.

| Problem | State Space Size | | Size of $\aleph$ | | Solution |
|---|---|---|---|---|---|
| | full | effective | max | aver. | time (s) |
| Coffee | 32 | 12 | 102 | 56 | 47 |
| Widget | 32 | 14 | 205 | 121 | 397 |
| Pavement | 128 | 85 | 39 | 16 | 250 |

Table 2: Solution statistics for the three test problems

Each of the three problems was solved using Hansen and Feng's [8] ADD implementation of incremental pruning (IP) to produce a set $\aleph$ of $\alpha$-vectors using a compact ADD representation. Each problem is run to 15 stages (discounted). Table 2 shows, for each problem, its full state space size, $|\mathcal{S}|$, and its *effective size*, the largest intersection of abstract (ADD) states encountered during solution (specifically, the LP-dominance test in IP). The effective size is more relevant to solution time than $|\mathcal{S}|$. We also show the solution time (in seconds) along with the average size of the sets $\aleph$ over the fifteen stages and the maximum size set.

Once solved, we searched for a good projection scheme for each POMDP by minimizing different error bounds and/or using different switch tests, as described above. Specifically, six algorithms are tested: the $B$-bound and $E$-bound search of [15], which computes switch sets using an LP and chooses a projection using either the $B$ or $E$ error bounds; the VS analogs of these procedures which computes weaker VS-switch sets using the algebraic formulation of Section 3.2; and the VS search methods (sum and max) of Section 3.3, which ignore these bounds, but instead try to minimize Eq. 2 or Eq. 3. All search algorithms perform a lattice search within the set of projection schemes that partition variables in disjoint subsets. Furthermore, assuming that marginals of at most two variables provide a suitable efficiency/accuracy tradeoff, the lattice is traversed until all children of a node correspond to projections with a marginal with 3 variables. This last node is the projection scheme returned by the search.

We compare the time required to find a good projection using the different search procedures in Table 3. As expected, the running time is much less when using VS-switch tests (compared to LP-switch tests), since VS-switch tests do not require the solution of LPs. As for VS search algorithms, whether we minimize the *sum* of the relative error rates or their *maximum*, the running time is roughly the same and it is significantly faster than $B$-bound and $E$-bound search algorithms that use LP-switch tests, but a bit slower if VS switch tests are used for $B$-bound search. This is because, on the one hand, the VS search does not solve LPs (compared to LP-switch tests), but on the other hand, it has a stronger dependence on the number of $\alpha$-vectors (compared to VS-switch tests). The time to search for good projections *can be* much worse than that of solving POMDPs (though this offline cost translates into online gains). In fact, only search procedures that avoid solving LPs scale effectively to larger problems. In some cases, these offer a decrease of up to two orders of magnitude. The running time of VS procedures is roughly of the same order of magnitude as that of the POMDP solution procedures.

| Problem | Solut. time | B-bd search LP | B-bd search VS | E-bd search LP | E-bd search VS | VS search max | VS search sum |
|---|---|---|---|---|---|---|---|
| Coffee | 47 | 1019 | 30 | 4379 | 2651 | 151 | 154 |
| widget | 397 | 10142 | 109 | 89605 | 48695 | 707 | 703 |
| Pavement | 250 | 345 | 35 | 841 | 126 | 97 | 96 |

Table 3: Search running time in seconds

| | Error | B-bd search LP | B-bd search VS | E-bd search LP | E-bd search VS | VS search max | VS search sum |
|---|---|---|---|---|---|---|---|
| Single Approx | Aver. | 0.0013 | 0.0063 | 0.0063 | 0.0063 | 0.0013 | 0.0014 |
| | B-bd | 3.2840 | 5.9150 | 4.3910 | 5.9150 | 3.2840 | 3.2840 |
| Several Approx | Aver. | 0.0144 | 0.0161 | 0.0161 | 0.0161 | 0.0154 | 0.0107 |
| | E-bd | 13.085 | 13.085 | 13.085 | 13.085 | 13.085 | 13.085 |

Table 4: Coffee problem: error comparisons

| | Error | B-bd search LP | B-bd search VS | E-bd search LP | E-bd search VS | VS search max | VS search sum |
|---|---|---|---|---|---|---|---|
| Single Approx | Aver. | 0.0352 | 0.0352 | 0.0352 | 0.0352 | 0.0082 | 0.0081 |
| | B-bd | 3.4080 | 3.6270 | 3.4080 | 3.6270 | 3.4080 | 3.4080 |
| Several Approx | Aver. | 0.0509 | 0.0508 | 0.0508 | 0.0508 | 0.0519 | 0.0517 |
| | E-bd | 8.3811 | 8.3811 | 8.3811 | 8.3811 | 8.3811 | 8.3811 |

Table 5: Widget problem: error comparisons

| | Error | B-bd search LP | B-bd search VS | E-bd search LP | E-bd search VS | VS search max | VS search sum |
|---|---|---|---|---|---|---|---|
| Single Approx | Aver. | 0.0015 | 0.0015 | 0.0015 | 0.0015 | 0.0014 | 0.0014 |
| | B-bd | 5.3860 | 5.6900 | 5.3860 | 5.6900 | 5.3680 | 5.6160 |
| Several Approx | Aver. | 0.0066 | 0.0066 | 0.0066 | 0.0066 | 0.0071 | 0.0028 |
| | E-bd | 23.218 | 35.392 | 23.498 | 35.392 | 23.874 | 24.384 |

Table 6: Pavement problem: error comparisons

We also compare the actual average error, as well as the formal $B$ and $E$ error bounds, obtained when applying the projection schemes found by various search algorithms (Tables 4, 5 and 6). The average error is the average loss incurred for 5000 random initial belief states generated from a uniform distribution. We see that the average error is essentially the same whether the VS search procedure is used or some error bound is minimized. As a result, the dramatic computational savings associated with the VS procedures has effectively no impact on solution quality. Note that the $B$ and $E$ bounds are much larger than the average error observed because the bounds are concerned with the worst case scenario and, furthermore, they are not tight (supersets of the switch sets are really computed).

## 5 Concluding Remarks

We have proposed a new approach to value-directed belief state approximation for POMDPs. Our vector space approach—using either VS-switch tests or direct VS search—offers significant computational benefits over the value-directed methods proposed by Poupart and Boutilier [15]. While the error bounds are looser, we have seen in practice that our new schemes perform as well as the others with respect to solution quality; thus the computational savings are achieved with little impact on decision quality. Furthermore, the vector space model provides new insights into the belief state approximation problem and how approximation impacts decision quality.

This novel view also gives us access to numerous tools from linear algebra to design approximation methods that could potentially offer better tradeoffs between decision quality and monitoring efficiency. For instance, it would be interesting to investigate linear projectors since they allow the design of linear approximation methods by specifying (among other things) a displacement subspace $D_S$ which could be made as perpendicular as possible to the gradient vectors $\alpha_{ij}$. Linear projectors are well-studied approximation methods with numerous properties and therefore they provide a promising alternative for improving value-directed approximate belief state monitoring.

The success and scalability of our methods strongly depends on the structure and compactness of the $\alpha$-vectors. Therefore, one could also analyze the dependency between the $\alpha$-vector structure and the conditional independence structure of the transition and observation functions. From a linear algebra perspective, the $\alpha$-vectors can be viewed as a discounted sum of reward vectors multiplied by transition and observation matrices. Thus compact and structured $\alpha$-vectors could arise when the reward vectors fall into a small invariant subspace of the transition and observation matrices. A possible direction of research would then be to relate the conditional independence structure of the transition and observation functions with their eigenvalue and eigenvector properties since they define the invariant subspaces. This would allow us to better characterize the situations in which our approach is suitable.

We are currently extending this approach, and its analysis,

in a number of different directions. First, we motivated this work by focusing on infinite-horizon POMDPs, though our algorithms and analysis assume a finite set of $\alpha$-vectors. Often one is forced to approximate the value function (e.g., by producing a finite set of vectors where an infinite set is required, or simply by reducing the number of vectors to keep it manageable in size). Our algorithms can be applied directly to approximate value functions, and we expect that the analysis can be extended with suitable modifications as well. We are also interested in applying the idea of value-directed monitoring to other representations of value functions and other forms of approximate monitoring. The use of grid-based value functions [4, 9, 10] provides a very attractive method for producing approximate value functions for which approximate monitoring will generally be necessary. We expect that information in grid-based value functions can be used profitably to direct the choice of projection (or other approximation) schemes. The use of value information to guide other belief state approximation methods is also of tremendous interest: we have recently developed a sampling (particle filtering) algorithm that is influenced by value function information [16]. Finally, if it is taken for granted that some form of belief state approximation will be used, one might attempt to solve the POMDP to account for this fact; that is, can we construct policies that are optimal subject to the resource constraints placed on the monitoring process?

**Acknowledgements:** This research was supported by the Natural Sciences and Engineering Research Council and the Institute for Robotics and Intelligent Systems.